\documentclass[conference]{IEEEtran}
\IEEEoverridecommandlockouts
\usepackage{cite}
\usepackage{amsmath,amssymb,amsfonts}
\usepackage{algorithmic}
\usepackage{graphicx}
\usepackage{textcomp}
\usepackage{xcolor}
\def\BibTeX{{\rm B\kern-.05em{\sc i\kern-.025em b}\kern-.08em
    T\kern-.1667em\lower.7ex\hbox{E}\kern-.125emX}}

\usepackage{booktabs}
\usepackage{nicefrac}
\usepackage{microtype}
\usepackage{multirow}
\usepackage{enumitem}
\usepackage{xspace}
\usepackage{pifont}
\usepackage[utf8]{inputenc}
\usepackage[T1]{fontenc}
\usepackage{hyperref}
\usepackage{url}

\let\citep\cite
\let\citet\cite

\newcommand{\method}{{MuteBench}\xspace}

\begin{document}
\title{\method: Modality Unavailability Tolerance Evaluation for Incomplete Multimodal Fusion}
\author{%
\IEEEauthorblockN{Wugeng Zheng\textsuperscript{\textdagger}}
\IEEEauthorblockA{University of Central Florida\\
Orlando, FL, USA\\
\texttt{wgzheng@ucf.edu}}
\and
\IEEEauthorblockN{Ziwen Kan}
\IEEEauthorblockA{University of Central Florida\\
Orlando, FL, USA\\
\texttt{zi605672@ucf.edu}}
\and
\IEEEauthorblockN{Tianyu Hu}
\IEEEauthorblockA{University of Central Florida\\
Orlando, FL, USA\\
\texttt{ti721494@ucf.edu}}
\and[\hfill\mbox{}\par\vspace{1em}\mbox{}\hfill]
\IEEEauthorblockN{Tianlong Chen}
\IEEEauthorblockA{University of North Carolina\\
at Chapel Hill\\
Chapel Hill, NC, USA\\
\texttt{tianlong@cs.unc.edu}}
\and
\IEEEauthorblockN{Chen Chen}
\IEEEauthorblockA{University of Central Florida\\
Orlando, FL, USA\\
\texttt{chen.chen@crcv.ucf.edu}}
\and
\IEEEauthorblockN{Song Wang\textsuperscript{*}}
\IEEEauthorblockA{University of Central Florida\\
Orlando, FL, USA\\
\texttt{song.wang@ucf.edu}}
\thanks{\textsuperscript{\textdagger}First author. \textsuperscript{*}Corresponding author.}
}

\maketitle

\begin{abstract}
\label{abstract}
Multimodal physiological data supports clinical AI from intensive care to wearable monitoring, yet incomplete inputs are common. We study two forms of missingness: whole-channel or stream loss and contiguous gaps within an active signal. We present \method, a benchmark covering 9 datasets from 6 clinical and physiological domains, 6 fusion architectures, and 2 controlled missingness modes over 390,000 samples. Across the benchmark, robustness varies with fusion design, data structure, and training exposure. In this benchmark, the evaluated MoE-fusion group shows the largest AUROC degradation in 16 of 18 modality-missing dataset--rate cells. Modality missing causes a larger AUROC drop than within-modality missing on seven of nine datasets. A PTB-XL case study shows that diffusion-based imputation improves AUROC under within-modality missing and lowers AUROC under modality missing. These results provide a reproducible comparison of multimodal fusion systems under incomplete physiological data.
\end{abstract}

\begin{IEEEkeywords}
Multimodal fusion, missing data, physiological time series, robustness evaluation, benchmarking.
\end{IEEEkeywords}

% -------------------------------------------------------
\section{Introduction}
\label{sec:intro}

Multimodal physiological data is central to modern clinical AI \citep{acosta2022multimodal, soenksen2022integrated}, combining ECG, EEG, PPG, and structured records to support disease prediction, severity assessment, and health monitoring \citep{warner2024multimodal, shickel2019deepsofa, dunn2021wearable}. Yet sensors routinely fail in practice. Leads detach, devices lose contact, and transmission errors drop segments of data \citep{chromik2022alarm, vila2021ecg}, resulting in two distinct failure modes. \emph{Modality missing} occurs when an entire channel is absent \citep{zhang2022m3care}, while \emph{within-modality missing} refers to the loss of a contiguous time segment within an active channel \citep{che2018grud}. Both failures arise during deployment and data collection alike, making incomplete multimodal data a common feature of clinical datasets.

Despite this, most multimodal models are trained and evaluated assuming complete modality availability \citep{WANG2024e26772, finlayson2021clinician}, overlooking the incompleteness common in clinical practice. Even when data are complete, training dynamics tend to favor certain modalities over others \citep{9156420, pmlr-v162-huang22e}, so many multimodal models in practice rely on one dominant modality while leaving others underused. Missing modalities therefore worsen these existing imbalances rather than creating an entirely new challenge. Recent work on ICU time-series imputation reports different method rankings across random, contiguous-gap, and sensor-specific missingness \citep{icumissing2026}. Few benchmarks jointly evaluate multiple architectures under whole-channel and within-channel missingness at controlled severity levels across diverse clinical signals (Table~\ref{tab:benchmark-comparison}).

To address this gap, we introduce the Modality Unavailability Tolerance Evaluation Benchmark (\method), covering 9 datasets from 6 clinical domains, over 390,000 total samples across datasets ranging from 559 to 277,179 samples, and 6 fusion models from three architecture families: channel-independent models, Mixture-of-Experts fusion models, and shared-specific decomposition models. We evaluate all models under two controlled missingness conditions at two severity levels, yielding 270 evaluation configurations replicated across three independent random seeds (810 total runs). The datasets span channel counts from 2 to 78, sequence lengths from 48 to 3,000 time steps, and sampling rates from once-per-hour clinical observations to 4,000\,Hz physiological waveforms. A framework-agnostic missingness library applies deterministic masks consistently across codebases, reducing variation from data pipelines. Our main contributions are:
\begin{itemize}[leftmargin=1.5em, topsep=2pt, itemsep=1pt]
    \item \textbf{A controlled benchmark for incomplete multimodal physiological data.} \method crosses 9 datasets from 6 clinical domains, 6 fusion architectures from three model families, and two missingness modes at controlled severity levels under a shared evaluation protocol.
    \item \textbf{Empirical evidence about robustness factors.} Across the benchmark and capacity-controlled analyses, observed rankings depend on fusion design and dataset structure; parameter count alone does not explain them. Channel-independent and shared-specific models show different strengths across missingness modes, while curriculum dropout helps within its training range.
    \item \textbf{A focused imputation case study.} On PTB-XL, diffusion-based imputation improves downstream classification under within-modality missing for the evaluated models, while providing little benefit when an entire input channel is absent. This result defines a useful case study rather than a general conclusion about imputation.
\end{itemize}

\begin{table*}[!t]
\centering
\small
\setlength{\tabcolsep}{3.9pt}
\caption{Feature comparison of multimodal robustness benchmarks. \method evaluates multiple fusion architectures under two controlled missingness conditions across diverse clinical physiological datasets.}
\label{tab:benchmark-comparison}
\begin{tabular}{lcccccccc}
\toprule[1pt]
\textbf{Benchmark} &
\textbf{Clinical} &
\shortstack{\textbf{Physio.}\\\textbf{TS}} &
\shortstack{\textbf{Multi-}\\\textbf{dataset}} &
\shortstack{\textbf{Multi-}\\\textbf{arch}} &
\shortstack{\textbf{Mod.}\\\textbf{Missing}} &
\shortstack{\textbf{Within-mod.}\\\textbf{Missing}} &
\shortstack{\textbf{Severity}\\\textbf{Levels}} &
\shortstack{\textbf{Public}\\\textbf{Protocol}} \\
\midrule
MultiBench \citep{liang2023multibench}   & \ding{55} & \ding{55} & \ding{51} & \ding{51} & \ding{51} & \ding{55} & \ding{55} & \ding{51} \\
BenchMD \citep{wantlin2023benchmd}       & \ding{51} & \ding{51} & \ding{51} & \ding{55} & \ding{55} & \ding{55} & \ding{55} & \ding{51} \\
MC-BEC \citep{Chen2023MultimodalCB}      & \ding{51} & \ding{55} & \ding{51} & \ding{55} & \ding{55} & \ding{55} & \ding{55} & \ding{55} \\
CLIMB \citep{dai2025climb}               & \ding{51} & \ding{51} & \ding{51} & \ding{55} & \ding{55} & \ding{55} & \ding{55} & \ding{51} \\
M3Care \citep{zhang2022m3care}           & \ding{51} & \ding{51} & \ding{55} & \ding{55} & \ding{51} & \ding{55} & \ding{51} & \ding{51} \\
\midrule
\textbf{MuteBench (Ours)}               & \ding{51} & \ding{51} & \ding{51} & \ding{51} & \ding{51} & \ding{51} & \ding{51} & \ding{51} \\
\toprule[1pt]
\end{tabular}
\end{table*}

These results give \method a practical role: it lets researchers compare fusion mechanisms under the same controlled missingness conditions and see how dataset structure changes the ranking. Channel count, sequence length, and modality alignment are useful design variables when developing models for incomplete-on-arrival data. The benchmark therefore offers a common evaluation frame for studying graceful degradation in clinical multimodal fusion.

\section{Related Work}
\label{sec:related_work}

\textbf{Clinical Multimodal Benchmarks.}
\label{clinical_multimodal_benchmarks}
Clinical multimodal benchmarks fall into two categories. The first covers medical question answering~\citep{app11146421, 53083}, LLM and LVLM reasoning on clinical images and guidelines~\citep{tu2024medpalmm, 10655261, li2025medguide, fast2024amega, jiang2026m3cotbench}, and multimodal QA over EHR tables paired with radiology images~\citep{NEURIPS2023_0c007ebe}. These all focus on text-based or knowledge-driven reasoning without continuously sampled physiological signals. The second targets task-oriented clinical prediction by combining time series, waveforms, and structured features~\citep{Chen2023MultimodalCB, zhang2022m3care, wantlin2023benchmd}. Liang et al.~\cite{liang2023multibench} introduce a general-purpose benchmark with robustness evaluation but do not separate modality missing from within-modality missing patterns and do not focus on clinical physiological signals. Existing benchmarks leave open how different missing patterns interact with signal structure across multiple fusion architectures and clinical physiological datasets.

\textbf{Missing Data in Clinical Time Series.}
\label{missing_data_in_clinical_ts}
Most multimodal healthcare prediction models are developed and benchmarked assuming complete modality availability~\citep{WANG2024e26772}, yet this assumption rarely holds in real clinical practice.
Missing data in clinical physiological signals takes two forms: modality missing (an entire sensor stream absent) and within-modality missing (contiguous temporal gaps within an active channel). Single-modality methods address these challenges through missingness-aware recurrent models~\citep{che2018grud}, continuous-time attention networks~\citep{shukla2021mtan}, and graph-based sensor-dropout encoders~\citep{zhang2022raindrop}. For multimodal settings, existing methods include generative reconstruction of absent modalities~\citep{ma2021smil}, shared-specific feature decomposition~\citep{wang2023multi, yao2024drfuse}, and prompt-based adaptation~\citep{lee2023map}. These methods are each proposed and evaluated in isolation on individual datasets, without a unified comparison across architectures or clinical signal structures.

\textbf{Benchmark for Robustness.}
\label{benchmark_for_robust}
Two gaps remain. First, few benchmarks jointly evaluate multiple fusion architectures under both missing patterns at controlled severity levels on clinical physiological signals. MultiBench~\citep{liang2023multibench} and MC-BEC~\citep{Chen2023MultimodalCB} each address only one of these dimensions. Second, no benchmark accounts for how dataset structure (modality alignment type, channel count, and sequence length) interacts with missing pattern type to determine degradation. Without this, it is unclear which architectural choices suit which deployment scenario. We address both gaps with nine clinical datasets across three modality alignment structures, six fusion architectures, and two missing patterns at two severity levels. Table~\ref{tab:benchmark-comparison} details this comparison. As Section~\ref{sec:exp} shows, architecture rankings can reverse between the two failure modes, making isolated evaluation misleading.

\section{Dataset}
\label{sec:dataset}

\begin{figure*}[!t]
    \centering
    \includegraphics[width=\linewidth]{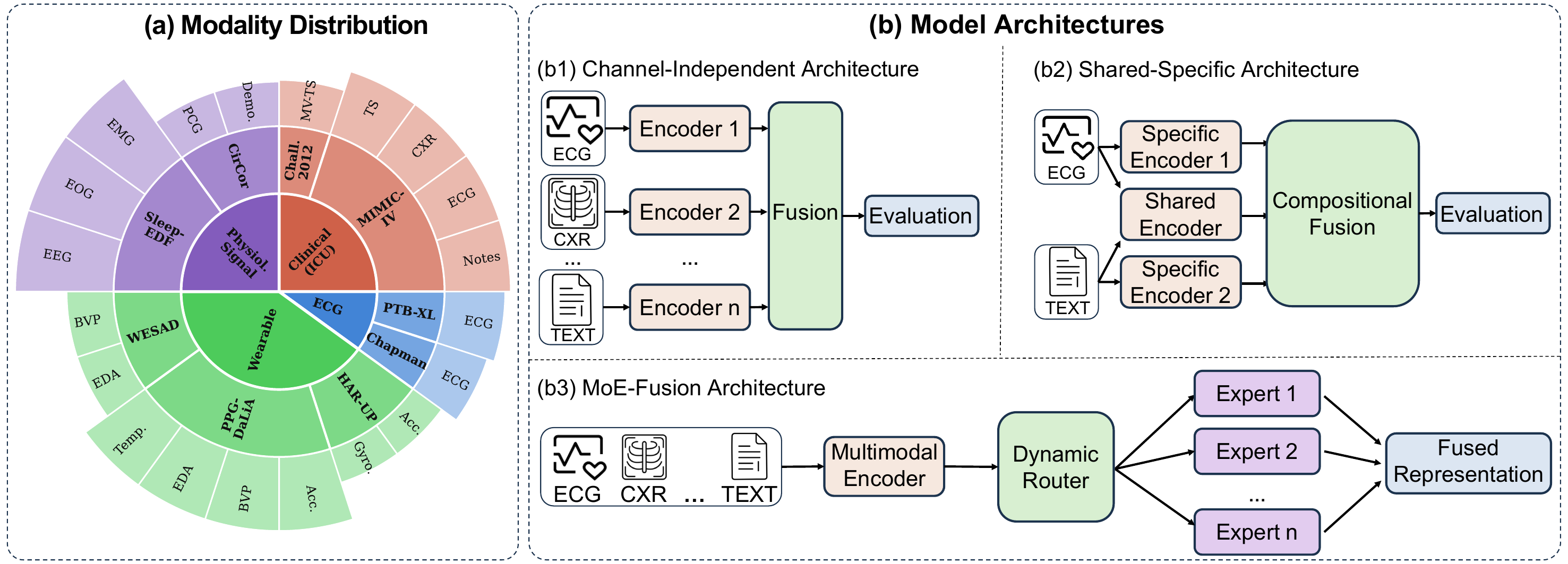}
    \caption{Overview of \method. We evaluate 9 datasets spanning 6 clinical domains and three representative fusion architecture families: channel-independent models, Mixture-of-Experts fusion models, and shared-specific decomposition models, under two systematic missingness patterns at controlled severity levels.}
    \label{fig:overall}
\end{figure*}

This section describes the datasets used in \method. We select 9 datasets from 6 clinical domains (Figure~\ref{fig:overall}, left), covering varied clinical and physiological settings so that the benchmark spans different signal structures and task formats. The following subsections describe our selection criteria and modality definitions.

\subsection{Dataset Selection and Modality Definitions}
\label{subsec:data_selection}
We curate 9 datasets spanning 6 domains (e.g., ICU, cardiology, wearable activity tracking) and multiple task formats (binary, multi-class, multi-label) to measure both spatial and temporal robustness. This diversity ensures that the evaluated fusion strategies are tested across varied input structures and clinical objectives, enabling a thorough assessment of generalization across diverse clinical settings.

We categorize the datasets into three modality alignment types to isolate how structural alignment affects a model's ability to recover missing information. \textbf{Type 1 (Homogeneous and Aligned)} datasets share the same format and time axis (e.g., 12-lead ECG), testing pure spatial compensation. \textbf{Type 2 (Heterogeneous and Aligned)} datasets have different physical domains but synchronized time axes (e.g., EEG and respiration), testing cross-domain temporal reasoning. \textbf{Type 3 (Heterogeneous and Unaligned)} datasets lack both spatial and temporal synchronization (e.g., asynchronous clinical records combined with static metadata), testing models under the most challenging alignment conditions. The dataset definitions and alignment types are specified in this section.

\begin{figure*}[!t]
    \centering
    \includegraphics[width=\linewidth]{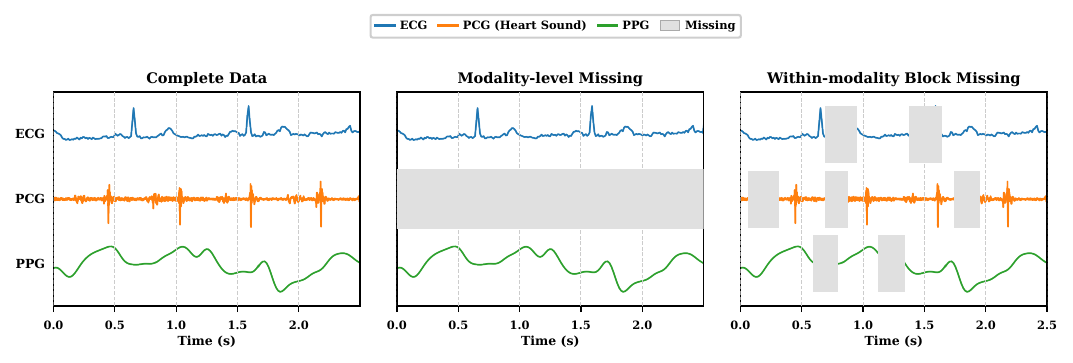}
    \caption{Evaluation of missing-data conditions.
    \textbf{Left (Complete):} Original, fully observed signals.
    \textbf{Middle (Modality missing):} Entire modalities (e.g., B and E) are dropped with probability $p$, simulating whole-sensor failures.
    \textbf{Right (Within-modality missing):} Contiguous time segments are masked independently per channel, simulating transient interruptions like motion artifacts.}
    \label{fig:missing_types}
\end{figure*}

\subsection{Dataset Statistics}
\label{subsec:dataset_stats}
The benchmark spans 9 datasets from 6 clinical and physiological domains, totalling over 390,000 samples. Dataset sizes range from HAR-UP \citep{martinezvillasenor2019upfall} with 559 samples to Sleep-EDF \citep{kemp2000sleepedf} with 277,179 epochs. ICU datasets fall in the mid-range, with MIMIC-IV \citep{johnson2023mimiciv} contributing 5,100 patient episodes and Challenge-2012 \citep{silva2012challenge20126} about 4,000 hospital stays. The cardiac datasets include PTB-XL \citep{PhysioNet-ptb-xl-1.0.1} with 21,837 recordings and Chapman-Shaoxing \citep{zheng2020chapman} with 10,247 recordings. Task formats are equally diverse. Three datasets target binary classification: HAR-UP, MIMIC-IV, and Challenge-2012. Four require multi-class prediction with 3 to 9 categories: Sleep-EDF distinguishes 5 sleep stages, PPG-DaLiA covers 9 activity types, WESAD \citep{schmidt2018wesad} recognizes 3 affective states, and CirCor \citep{oliveira2022circor} uses absent, unknown, and present murmur labels. The remaining two, PTB-XL and Chapman-Shaoxing, use multi-label prediction with 5 and 7 diagnostic label groups per recording. All datasets are evaluated with Macro-AUROC as the primary metric and Macro-F1 reported alongside.

The datasets also vary substantially in input structure. Channel counts range from 2 in WESAD (wrist EDA and BVP sensors) to 78 in CirCor (64 mel-spectrogram bins plus 14 static demographic channels). Sequence lengths span from 48 time steps in MIMIC-IV and Challenge-2012 (hourly ICU bins) to 3,000 steps in Sleep-EDF (30-second windows at 100\,Hz). Other examples include PTB-XL and Chapman-Shaoxing with 12 ECG leads at $T{=}250$ and $T{=}1{,}000$, PPG-DaLiA with 9 wearable channels at $T{=}256$, and HAR-UP with 30 IMU channels from 5 body-worn sensors at $T{=}140$. Sampling rates vary from once-per-hour clinical observations to 4,000\,Hz phonocardiogram waveforms. Challenge-2012 has extreme temporal sparsity from irregular clinical measurements, making it the most irregularly sampled dataset in the benchmark. This structural variation is deliberate: it tests models across low- and high-channel regimes, short- and long-horizon reasoning, and all three modality alignment types, from homogeneous synchronized signals like PTB-XL and Chapman-Shaoxing to heterogeneous unaligned mixtures of time series and static embeddings in MIMIC-IV and CirCor. Table~\ref{tab:dataset_overview} summarizes the per-dataset statistics.

\section{Experiments}
\label{sec:exp}

\subsection{Experimental Setup}
\label{subsec:exp_setup}

This section presents the experimental results and analysis. We evaluate model robustness against incomplete data across diverse settings, guided by three questions: \textbf{(a) Datasets}: Do models show consistent robustness across dataset types and clinical domains? \textbf{(b) Models}: Can different fusion architectures maintain robustness under missing data? \textbf{(c) Missingness}: Do models respond differently to modality missing versus within-modality missing? The datasets selected and model configurations used throughout this work are described as follows.

\paragraph{Datasets:} We use 9 datasets covering the domains and modality types defined in Section \ref{sec:dataset}. These include medical data (MIMIC-IV, Challenge-2012), heart signals (PTB-XL, Chapman-Shaoxing), cardiac auscultation (CirCor), brain signals (Sleep-EDF), wearable data (PPG-DaLiA, WESAD), and activity tracking (HAR-UP). The benchmark includes 18,885 PTB-XL patients, 52 Sleep-EDF subjects, 15 PPG-DaLiA subjects, 17 HAR-UP subjects, and 942 CirCor patients; the remaining datasets are described by their admission or recording units in Table~\ref{tab:dataset_overview}. Categorized by modality structure, these datasets span three types: Type~1 (homogeneous and aligned), Type~2 (heterogeneous and aligned), and Type~3 (heterogeneous and unaligned), as defined in this section. Table~\ref{tab:dataset_overview} lists the specific details for all 9 datasets.

\begin{table*}[!t]
\centering
		\renewcommand{\arraystretch}{1.1}
    \small
    \setlength\tabcolsep{7.5pt}
\caption{Overview of the specific datasets evaluated in the benchmark. The table is sorted by modality type and task type.}
\label{tab:dataset_overview}
\begin{tabular}{llccccccc}
\toprule[1.pt]
Dataset & Domain & Samples & Channels & Steps & Modality & Task Type & Metric \\ \hline
% Type 1 Group
HAR-UP & HAR & 559 & 30 & 140 & Type 1 & Binary & AUROC \\
PTB-XL & Cardio. & 21.8K & 12 & 250 & Type 1 & Multi-label & AUROC \\
Chapman & Cardio. & 10.2K & 12 & 1000 & Type 1 & Multi-label & AUROC \\ \hline
% Type 2 Group
Sleep-EDF & Neuro. & 277.2K & 5 & 3000 & Type 2 & Multi-class & AUROC \\
PPG-DaLiA & HAR & 64.7K & 9 & 256 & Type 2 & Multi-class & AUROC \\
WESAD & Affect & 4.4K & 2 & 480 & Type 2 & Multi-class & AUROC \\ \hline
% Type 3 Group
MIMIC-IV & ICU & 5.1K & 30 & 48 & Type 3 & Binary & AUROC \\
Challenge-2012 & ICU & 4K & 42 & 48 & Type 3 & Binary & AUROC \\
CirCor & Cardiac & 3.1K & 78 & 625 & Type 3 & Multi-class & AUROC \\
\toprule[1.pt]
\end{tabular}
\end{table*}

\begin{table*}[!t]
\centering
		\renewcommand{\arraystretch}{1.1}
    \small
    \setlength\tabcolsep{10pt}% Adjust column spacing
\caption{Summary of baseline model architectures and fusion strategies.}
\label{tab:model_summary}
\begin{tabular}{lllc}
\toprule[1.pt]
Model & Backbone Encoder & Key Fusion Strategy & Sparse MoE \\ \hline
CLIMB & Transformer & Channel-independent & No \\
Flex-MoE & Transformer & G-/S-Router + Modality Bank & Yes \\
MIRA & Neural ODE & Frequency-domain MoE & Yes \\
ShaSpec & CNN/MLP & Shared-specific Decomposition & No \\
FuseMoE & Transformer & Sparse MoE + mTAND & Yes \\
Maestro & Sparse Attention (per-modal) & Sparse Attn + Curriculum Dropout & Yes \\
\toprule[1.pt]
\end{tabular}
\end{table*}

\paragraph{Models:} We evaluate 6 representative multimodal fusion architectures spanning multiple design paradigms (Figure~\ref{fig:overall}, right): channel-independent models (CLIMB~\citep{dai2025climb}, MIRA~\citep{li2025mira}), a shared-specific decomposition model (ShaSpec~\citep{wang2023multi}), and MoE-fusion models (Flex-MoE~\citep{yun2024flexmoemodelingarbitrarymodality}, FuseMoE~\citep{han2024fusemoe}, Maestro~\citep{maestro2025}). Table~\ref{tab:model_summary} summarizes their architectures and key fusion strategies; all models use their published fusion modules and task-specific classification heads, MIRA is trained from scratch, checkpoints are selected by validation Macro-AUROC, and each configuration is run with three independent seeds.

\paragraph{Missing Data and Evaluation Protocol:} We evaluate two controlled stress conditions (Figure~\ref{fig:missing_types}) in an incomplete-on-arrival setting, where the same missingness distribution is present during development and evaluation. Under \textbf{modality missing}, the dropping unit follows the dataset representation: channel-stacked datasets drop channels, while MIMIC-IV follows its four semantic input streams. At least one input stream is always retained. Under \textbf{within-modality missing}, non-overlapping contiguous blocks (5--10\% of $T$ each) are masked per time-series channel to a total fraction $b \in \{0.2, 0.5\}$; static CirCor demographic channels are excluded. The two severity levels are controlled moderate and severe evaluation points. All patterns are generated deterministically from a \texttt{(seed, sample\_id)} pair and applied uniformly to train, validation, and test splits, so every model receives identical masks. Maestro ramps training dropout from zero to 40\% over the first 10 epochs; inference uses the shared evaluation masks without internal dropout.

\paragraph{Masking Units and Input Scope:} The masking unit follows each dataset's input representation. For PTB-XL, each ECG lead is treated as one input channel, so modality missing is implemented as lead-level channel masking. For CirCor, the evaluated channels are mel-spectrogram bins, while static demographic features are excluded from within-modality masking. MIMIC-IV uses four semantic input streams; its image, ECG, and text inputs are pre-extracted embeddings. The benchmark therefore evaluates robustness of the fusion layer rather than end-to-end image or language encoding.

\paragraph{Implementation and Training:} All six model projects share dataset splits and evaluation metrics while retaining their task-specific input adapters and training implementations. Shared index files and cached preprocessed arrays provide a common data interface. Missingness is injected on the fly in the dataset loader, allowing the same cached data to support all evaluated rates. Each sample contains the signal tensor $x\in\mathbb{R}^{C\times T}$, a per-timestep validity mask, a per-channel presence mask, and flags identifying time-series channels. Modality missing updates the channel-presence mask; within-modality missing updates the time-step mask and skips static channels. Multi-class tasks use inverse-frequency class weights, binary tasks use positive-class weighting, and multi-label tasks use per-class weights $w_{\mathrm{pos}}=n_{\mathrm{neg}}/n_{\mathrm{pos}}$.

\paragraph{Shared Evaluation Pipeline:} All six codebases use shared dataset splits, cached preprocessing outputs, and the same on-the-fly missingness injector. For each $(\texttt{seed},\texttt{sample\_id})$ pair, every model receives identical channel and block masks. Checkpoints are selected by validation Macro-AUROC, and class-imbalance weighting follows the task type.

\paragraph{Block Placement:} For a time-series channel of length $T$, block lengths range from $\ell_{\min}=\lceil0.05T\rceil$ to $\ell_{\max}=\lceil0.10T\rceil$. At target missing fraction $b$, the injector estimates the number of blocks as
\[
 k=\operatorname{round}\!\left(\frac{bT}{(\ell_{\min}+\ell_{\max})/2}\right).
\]
Start positions are sampled uniformly. Overlapping candidates are resampled up to 64 times; placement stops for that channel if no valid position is found. The sample-level generator draws a separate 64-bit seed for each channel before placement. Each channel then uses its own generator, producing different missing windows while preserving reproducibility for each $(\texttt{seed},\texttt{sample\_id})$ pair.

\subsection{Main Results}
\label{subsec:main_results}
We evaluate six models on nine datasets. We report AUROC and Macro-F1 under three conditions: clean data, modality missing, and within-modality missing. The tables report these as AUC and F1 columns for each model (AUC: AUROC; F1: Macro-F1).

\paragraph{Clean Data Results.} Table~\ref{tab:clean_data} shows clean-data baselines. No single model dominates: ShaSpec leads on Sleep-EDF, PPG-DaLiA, and Chapman; Maestro on PTB-XL, WESAD, HAR-UP, Challenge-2012, and CirCor; FuseMoE leads on MIMIC-IV (0.811 AUROC). These clean-data scores serve as the no-missing baseline shared by both modality-missing and within-modality-missing evaluations.

\begin{table*}[!t]
\centering
		\renewcommand{\arraystretch}{1.1}
    \small
\setlength{\tabcolsep}{5.5pt}% Adjust column spacing
\caption{Clean data results on nine datasets, averaged over three independent seeds. \textbf{AUC}: AUROC; \textbf{F1}: Macro-F1. Bold denotes the best result per dataset.}
\label{tab:clean_data}
\begin{tabular}{l | cc | cc | cc | cc | cc | cc}
\toprule[1.pt]
\multirow{2}{*}{\textbf{Dataset}} & \multicolumn{2}{c|}{\textbf{CLIMB}} & \multicolumn{2}{c|}{\textbf{MIRA}} & \multicolumn{2}{c|}{\textbf{Flex-MoE}} & \multicolumn{2}{c|}{\textbf{ShaSpec}} & \multicolumn{2}{c|}{\textbf{FuseMoE}} & \multicolumn{2}{c}{\textbf{Maestro}} \\
 & AUC & F1 & AUC & F1 & AUC & F1 & AUC & F1 & AUC & F1 & AUC & F1 \\
\hline
HAR-UP & .989 & .919 & .976 & .878 & .855 & .780 & .914 & .885 & .937 & .825 & \textbf{.991} & \textbf{.944} \\
PTB-XL & .838 & .545 & .830 & .496 & .772 & .466 & .865 & \textbf{.635} & .677 & .210 & \textbf{.883} & .621 \\
Chapman & .786 & .329 & .827 & .388 & .727 & .363 & \textbf{.850} & \textbf{.502} & .669 & .331 & .776 & .428 \\\hline
Sleep-EDF & .980 & .728 & .949 & .608 & .973 & .701 & \textbf{.984} & \textbf{.763} & .932 & .586 & .971 & .698 \\
PPG-DaLiA & .942 & .667 & .937 & .668 & .930 & .631 & \textbf{.959} & .735 & .927 & .627 & .956 & \textbf{.762} \\
WESAD & .701 & \textbf{.501} & .685 & .470 & .684 & .434 & .674 & .451 & .583 & .369 & \textbf{.736} & .494 \\\hline
MIMIC-IV & .782 & .411 & .741 & .371 & .760 & .403 & .656 & .311 & \textbf{.811} & \textbf{.420} & .805 & .393 \\
Challenge & .697 & .322 & .742 & .383 & .767 & .393 & .610 & .122 & .806 & .423 & \textbf{.812} & \textbf{.428} \\
CirCor & .649 & .447 & .601 & .387 & .567 & .357 & .678 & .444 & .655 & .409 & \textbf{.796} & \textbf{.573} \\
\toprule[1.pt]
\end{tabular}

\end{table*}

\begin{table*}[!t]
\centering
		\renewcommand{\arraystretch}{1.1}
    \small
\setlength{\tabcolsep}{4.5pt}% Adjust column spacing
\caption{Modality missing results on nine datasets at 20\% and 50\% drop rates, averaged over three independent seeds. \textbf{AUC}: AUROC; \textbf{F1}: Macro-F1. Bold denotes the best result per dataset per rate.}
\label{tab:modality_missing}
\begin{tabular}{l c | cc | cc | cc | cc | cc | cc}
\toprule[1.pt]
\multirow{2}{*}{\textbf{Dataset}} & \multirow{2}{*}{\textbf{Rate}} & \multicolumn{2}{c|}{\textbf{CLIMB}} & \multicolumn{2}{c|}{\textbf{MIRA}} & \multicolumn{2}{c|}{\textbf{Flex-MoE}} & \multicolumn{2}{c|}{\textbf{ShaSpec}} & \multicolumn{2}{c|}{\textbf{FuseMoE}} & \multicolumn{2}{c}{\textbf{Maestro}} \\
 & & AUC & F1 & AUC & F1 & AUC & F1 & AUC & F1 & AUC & F1 & AUC & F1 \\
\hline
\multirow{2}{*}{HAR-UP} & 20\% & .985 & .932 & .966 & .888 & .811 & .720 & .907 & .887 & .882 & .789 & \textbf{.987} & \textbf{.935} \\
 & 50\% & \textbf{.978} & \textbf{.893} & .936 & .855 & .752 & .676 & .887 & .865 & .780 & .696 & .962 & .878 \\
\hline
\multirow{2}{*}{PTB-XL} & 20\% & .822 & .512 & .825 & .498 & .737 & .452 & .860 & \textbf{.630} & .649 & .270 & \textbf{.873} & .611 \\
 & 50\% & .803 & .496 & .813 & .485 & .707 & .461 & .846 & \textbf{.614} & .602 & .233 & \textbf{.848} & .567 \\
\hline
\multirow{2}{*}{Chapman} & 20\% & .783 & .324 & .831 & .414 & .711 & .348 & \textbf{.849} & \textbf{.502} & .638 & .337 & .763 & .415 \\
 & 50\% & .772 & .314 & .810 & .397 & .681 & .345 & \textbf{.844} & \textbf{.505} & .580 & .291 & .726 & .391 \\
\hline
\multirow{2}{*}{Sleep-EDF} & 20\% & .973 & .688 & .926 & .560 & .962 & .654 & \textbf{.980} & \textbf{.737} & .890 & .505 & .952 & .645 \\
 & 50\% & .947 & .607 & .870 & .456 & .923 & .567 & \textbf{.960} & \textbf{.663} & .797 & .383 & .889 & .523 \\
\hline
\multirow{2}{*}{PPG-DaLiA} & 20\% & .930 & .645 & .933 & .663 & .909 & .570 & \textbf{.953} & \textbf{.712} & .901 & .562 & .931 & .675 \\
 & 50\% & .904 & .553 & .912 & .589 & .876 & .479 & \textbf{.936} & \textbf{.645} & .870 & .455 & .864 & .501 \\
\hline
\multirow{2}{*}{WESAD} & 20\% & .666 & .454 & .637 & .429 & .682 & .431 & .642 & .434 & .567 & .360 & \textbf{.721} & \textbf{.474} \\
 & 50\% & .649 & .444 & .620 & .427 & .675 & .415 & .651 & .448 & .592 & .374 & \textbf{.696} & \textbf{.471} \\
\hline
\multirow{2}{*}{MIMIC-IV} & 20\% & .750 & .361 & .725 & .364 & .737 & .376 & .638 & .298 & \textbf{.777} & .385 & .777 & \textbf{.395} \\
 & 50\% & .708 & .340 & .681 & .338 & .696 & .341 & .615 & .272 & \textbf{.736} & .346 & .729 & \textbf{.360} \\
\hline
\multirow{2}{*}{Challenge} & 20\% & .640 & .266 & .724 & .322 & .752 & .383 & .645 & .000 & .692 & .331 & \textbf{.781} & \textbf{.392} \\
 & 50\% & .667 & .314 & .697 & .298 & \textbf{.717} & \textbf{.347} & .607 & .168 & .616 & .257 & .714 & .312 \\
\hline
\multirow{2}{*}{CirCor} & 20\% & .652 & .428 & .596 & .395 & .529 & .315 & .659 & .443 & .599 & .387 & \textbf{.783} & \textbf{.561} \\
 & 50\% & .645 & .433 & .560 & .373 & .539 & .337 & .677 & .438 & .606 & .409 & \textbf{.755} & \textbf{.508} \\
\toprule[1.pt]
\end{tabular}
\end{table*}

\paragraph{Modality Missing Results.} Table~\ref{tab:modality_missing} shows results at 20\% and 50\% drop rates. On Chapman at 50\% drop, CLIMB loses 0.014 AUROC and MIRA loses 0.017. Maestro remains competitive on several datasets. FuseMoE loses 0.075 AUROC on PTB-XL at 50\% drop. The three-seed results report the observed run-to-run variability, including unstable Challenge-2012 runs.

\paragraph{Within-Modality Missing Results.} Table~\ref{tab:block_missing} shows within-modality missing results. ShaSpec is most robust on PTB-XL under block missing: its AUROC drops 0.007 at the 50\% block rate. Flex-MoE loses at most 0.010 AUROC on MIMIC-IV across the two block rates. Maestro shows a large F1 decrease on Sleep-EDF at block 50\% (F1: .443 vs.\ .698 clean, $\Delta=-0.255$).

\begin{table*}[!t]
\centering
		\renewcommand{\arraystretch}{1.1}
    \small
\setlength{\tabcolsep}{4.5pt}% Adjust column spacing
\caption{Within-modality missing results on nine datasets at 20\% and 50\% block rates, averaged over three independent seeds. \textbf{AUC}: AUROC; \textbf{F1}: Macro-F1. Bold denotes the best result per dataset per rate.}
\label{tab:block_missing}
\begin{tabular}{l c | cc | cc | cc | cc | cc | cc}
\toprule[1.pt]
\multirow{2}{*}{\textbf{Dataset}} & \multirow{2}{*}{\textbf{Rate}} & \multicolumn{2}{c|}{\textbf{CLIMB}} & \multicolumn{2}{c|}{\textbf{MIRA}} & \multicolumn{2}{c|}{\textbf{Flex-MoE}} & \multicolumn{2}{c|}{\textbf{ShaSpec}} & \multicolumn{2}{c|}{\textbf{FuseMoE}} & \multicolumn{2}{c}{\textbf{Maestro}} \\
 & & AUC & F1 & AUC & F1 & AUC & F1 & AUC & F1 & AUC & F1 & AUC & F1 \\
\hline
\multirow{2}{*}{HAR-UP} & 20\% & .983 & .919 & .969 & .901 & .826 & .742 & .883 & .851 & .866 & .782 & \textbf{.991} & \textbf{.941} \\
 & 50\% & .940 & \textbf{.843} & .937 & .830 & .795 & .710 & .866 & .820 & .842 & .756 & \textbf{.971} & .802 \\
\hline
\multirow{2}{*}{PTB-XL} & 20\% & .822 & .519 & .836 & .521 & .736 & .429 & .863 & \textbf{.634} & .662 & .236 & \textbf{.877} & .620 \\
 & 50\% & .798 & .457 & .818 & .478 & .706 & .404 & \textbf{.858} & \textbf{.630} & .608 & .158 & .856 & .514 \\
\hline
\multirow{2}{*}{Chapman} & 20\% & .777 & .313 & .814 & .400 & .706 & .343 & \textbf{.842} & \textbf{.493} & .634 & .323 & .767 & .369 \\
 & 50\% & .741 & .246 & .809 & .371 & .669 & .322 & \textbf{.835} & \textbf{.487} & .599 & .298 & .730 & .295 \\
\hline
\multirow{2}{*}{Sleep-EDF} & 20\% & .979 & .733 & .942 & .594 & .967 & .675 & \textbf{.983} & \textbf{.747} & .914 & .555 & .956 & .625 \\
 & 50\% & .973 & .698 & .927 & .568 & .961 & .650 & \textbf{.981} & \textbf{.740} & .887 & .505 & .893 & .443 \\
\hline
\multirow{2}{*}{PPG-DaLiA} & 20\% & .934 & .641 & .936 & .671 & .918 & .599 & \textbf{.950} & .688 & .920 & .604 & .943 & \textbf{.701} \\
 & 50\% & .907 & .573 & .934 & \textbf{.651} & .905 & .555 & \textbf{.939} & .620 & .914 & .567 & .882 & .467 \\
\hline
\multirow{2}{*}{WESAD} & 20\% & .696 & \textbf{.483} & .664 & .454 & .699 & .439 & .659 & .429 & .589 & .374 & \textbf{.726} & .460 \\
 & 50\% & .656 & \textbf{.448} & .622 & .411 & .681 & .412 & .639 & .415 & .586 & .362 & \textbf{.683} & .413 \\
\hline
\multirow{2}{*}{MIMIC-IV} & 20\% & .774 & .404 & .757 & .390 & .759 & .398 & .653 & .309 & \textbf{.805} & \textbf{.417} & .798 & .407 \\
 & 50\% & .752 & .379 & .744 & .357 & .750 & .383 & .645 & .309 & .794 & .374 & \textbf{.796} & \textbf{.412} \\
\hline
\multirow{2}{*}{Challenge} & 20\% & .680 & .299 & .729 & .341 & .757 & .387 & .638 & .006 & .800 & .393 & \textbf{.814} & \textbf{.433} \\
 & 50\% & .674 & .312 & .715 & .357 & .742 & .355 & .628 & .138 & \textbf{.793} & \textbf{.413} & .790 & .393 \\
\hline
\multirow{2}{*}{CirCor} & 20\% & .659 & .442 & .625 & .437 & .571 & .344 & .646 & .389 & .630 & .399 & \textbf{.774} & \textbf{.548} \\
 & 50\% & .632 & .430 & .565 & .367 & .572 & .352 & .663 & .368 & .623 & .397 & \textbf{.759} & \textbf{.477} \\
\toprule[1.pt]
\end{tabular}
\end{table*}

\subsection{Controlled Comparisons}
\label{subsec:controlled_comparisons}

The main benchmark compares published architectures with different parameter counts and training mechanisms. We therefore add three complementary checks that clarify the scope of the robustness findings.

\paragraph{Capacity-matched control.}
We retrain one representative from each family in a narrow 7.5--12.5M parameter range on WESAD and PTB-XL. Under the 50\% missingness settings, ShaSpec retains the highest mean AUROC across the two datasets ($0.8027\pm0.0082$), followed by CLIMB ($0.7369\pm0.0047$) and Flex-MoE ($0.6992\pm0.0087$), and ranks first in all four missing settings. The ordering is consistent across the three seeds. The per-dataset values are shown in Table~\ref{tab:capacity_matched}.

\begin{table*}[!t]
\centering
\small
\begin{tabular}{llccc}
\toprule
Model & Dataset & Clean & Modality 50\% & Block 50\% \\
\midrule
CLIMB (9.51M) & WESAD & .707$\pm$.005 & .644$\pm$.020 & .656$\pm$.021 \\
CLIMB (9.51M) & PTB-XL & .858$\pm$.004 & .819$\pm$.002 & .829$\pm$.003 \\
ShaSpec (8.47/10.47M) & WESAD & .783$\pm$.002 & .716$\pm$.024 & .729$\pm$.020 \\
ShaSpec (8.47/10.47M) & PTB-XL & .900$\pm$.000 & .872$\pm$.001 & .894$\pm$.001 \\
Flex-MoE (9.27/9.51M) & WESAD & .676$\pm$.018 & .682$\pm$.018 & .688$\pm$.015 \\
Flex-MoE (9.27/9.51M) & PTB-XL & .763$\pm$.005 & .711$\pm$.007 & .716$\pm$.001 \\
\bottomrule
\end{tabular}
\caption{Capacity-matched control. Each representative is trained within a 7.5--12.5M parameter range for 50 epochs on WESAD and PTB-XL. Values are mean$\pm$std test macro-AUROC over three seeds.}
\label{tab:capacity_matched}
\end{table*}

\paragraph{Family and scale analysis.}
Across the 657 matched benchmark observations used for this analysis, parameter count alone is not associated with robustness (Spearman $\rho=-0.03$, $p=.96$). After controlling for dataset and missingness condition, adding architecture family to a model with parameter count increases $R^2$ by .070, whereas adding parameter count to a model with family increases it by .032. Removing FuseMoE leaves the scale contribution at .000.

Table~\ref{tab:scale_summary} summarizes each model across datasets and all four missingness conditions (two modes at two rates). Each change is paired with the clean result for the same model, dataset, and seed. Parameter count correlates with retained AUROC at Spearman $\rho=-0.60$, and at equal parameter count the channel-independent and shared-specific groups lose .017 and .020 fewer AUROC points than MoE-fusion models ($p<.001$).

\begin{table}[!t]
\centering
\small
\setlength{\tabcolsep}{3.5pt}
\caption{Model-level scale summary over 657 matched observations, pooled across datasets and missingness conditions. Clean and retained values are mean AUROC; $\Delta_{\mathrm{ret}}$ is retained minus clean. Negative values indicate a decrease.}
\label{tab:scale_summary}
\begin{tabular}{l r r r r}
\toprule
Model & Params & Clean & Retained & $\Delta_{\mathrm{ret}}$ \\
\midrule
ShaSpec & 2.3M & .799 & .787 & $-.012$ \\
MIRA & 30.0M & .798 & .776 & $-.022$ \\
CLIMB & 9.5M & .818 & .796 & $-.022$ \\
Flex-MoE & 1.5M & .782 & .753 & $-.028$ \\
Maestro & 1.0M & .856 & .825 & $-.031$ \\
FuseMoE & 67.6--805.7M & .777 & .729 & $-.049$ \\
\bottomrule
\end{tabular}
\end{table}

\paragraph{Cross-site transfer.}
We train on PTB-XL, freeze the backbone, fit a linear target head, and evaluate on Chapman-Shaoxing. CLIMB reaches .8316 clean, .7987 under modality missing, and .7925 under block missing. ShaSpec reaches .7664, .7526, and .7231, while Flex-MoE reaches .7285, .6584, and .6699.

\begin{table}[!t]
\centering
\small
\begin{tabular}{lccc}
\toprule
Model & Clean & Modality 50\% & Block 50\% \\
\midrule
CLIMB & .832$\pm$.001 & .799$\pm$.007 & .793$\pm$.015 \\
ShaSpec & .766$\pm$.006 & .753$\pm$.006 & .723$\pm$.005 \\
Flex-MoE & .729$\pm$.011 & .658$\pm$.024 & .670$\pm$.004 \\
\bottomrule
\end{tabular}
\caption{Cross-site transfer from PTB-XL to Chapman-Shaoxing. The backbone is frozen, the target head is trained on clean data, and missingness is injected only at test time. Values are mean$\pm$std test macro-AUROC over three seeds.}
\label{tab:cross_site}
\end{table}

\subsection{Insights}
\label{subsec:insights}
From the benchmark results across nine datasets and six fusion architectures, we identify three recurring patterns. Several Challenge-2012 runs show model collapse.

\paragraph{Fusion design and parameter count show different relationships with robustness.}
Under modality missing, MoE-fusion models show the largest AUROC degradation in 16 of 18 dataset--rate cells in the family summary (Table~\ref{tab:family_degradation}). At 50\%, their family-level drops are .089 on Sleep-EDF, .058 on PTB-XL, .061 on Chapman-Shaoxing, .072 on MIMIC-IV, .113 on Challenge-2012, .067 on PPG-DaLiA, .014 on WESAD, .096 on HAR-UP, and .040 on CirCor. The shared-specific and channel-independent groups generally degrade less. In the capacity-matched control, ShaSpec has the highest mean AUROC across WESAD and PTB-XL under 50\% missingness ($0.8027\pm0.0082$), followed by CLIMB ($0.7369\pm0.0047$) and Flex-MoE ($0.6992\pm0.0087$). The family--scale analysis reports Spearman $\rho=-0.03$ and incremental $R^2$ values of .070 for family and .032 for parameter count.

\paragraph{Missingness impact differs across datasets.}
Modality missing causes a larger AUROC drop than within-modality missing on seven of the nine datasets; Chapman-Shaoxing and WESAD are the exceptions. The per-dataset breakdowns are shown in Tables~\ref{tab:modality_missing} and~\ref{tab:block_missing}.

\paragraph{Curriculum dropout and evaluation rate.}
Maestro is trained with a maximum curriculum dropout of 40\%. On PPG-DaLiA, the 50\% evaluation setting removes about 4.5 of the 9 channels on average. Maestro's AUROC is .956 on clean inputs and .864 at 50\% modality missing.

\begin{figure*}[t]
    \centering
    \includegraphics[width=\linewidth]{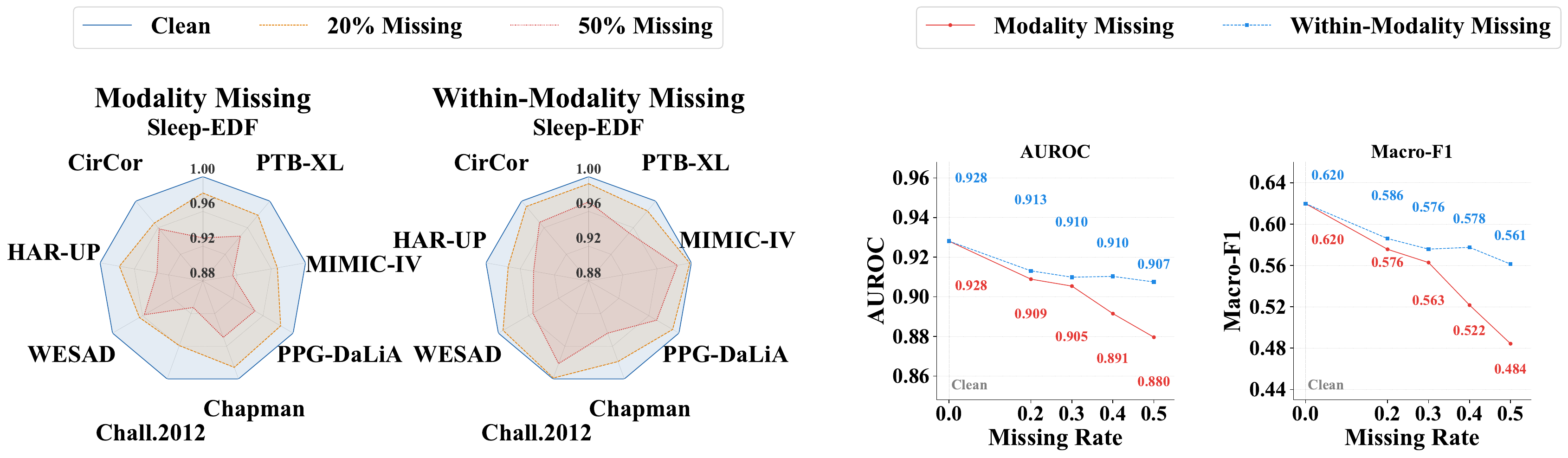}
    \caption{Degradation analysis. \textbf{left:} Radar chart of AUROC drop ($\Delta$AUROC $=$ clean $-$ missing, averaged over three seeds) across all 9 datasets under modality and within-modality missing at 20\% and 50\% rates; larger area indicates greater overall sensitivity, and each axis corresponds to one dataset. \textbf{right:} Detailed degradation trajectory of Flex-MoE on PPG-DaLiA: both AUROC and Macro-F1 decline steeply as missing rate increases, with modality missing inducing sharper drops than within-modality missing.}
    \label{fig:degradation}
\end{figure*}

\begin{table*}[!t]
\centering
		\renewcommand{\arraystretch}{1.1}
    \small
\setlength{\tabcolsep}{7pt}% Adjust column spacing
\caption{Effect of diffusion-based imputation on PTB-XL under modality missing. \textbf{Raw}: models trained on zero-filled dropped-modality input. \textbf{Diffusion}: same model trained on diffusion-reconstructed input. $\Delta$ denotes the performance change from Raw to Diffusion Imputed. \textbf{AUC}: AUROC; \textbf{F1}: Macro-F1. Three seeds; mean $\pm$ std.}
\label{tab:csdi_modality}
\begin{tabular}{l c | cc | cc | cc}
\toprule[1.pt]
\multirow{2}{*}{\textbf{Model}} & \multirow{2}{*}{\textbf{Rate}} &
  \multicolumn{2}{c|}{\textbf{Raw}} &
  \multicolumn{2}{c|}{\textbf{Diffusion Imputed}} &
  \multicolumn{2}{c}{\textbf{$\Delta$}} \\
 & & AUC & F1 & AUC & F1 & AUC & F1 \\
\hline
\multirow{2}{*}{CLIMB}    & 20\% & .822 $\pm$ .009 & .512 $\pm$ .012 & .816 $\pm$ .001 & .503 $\pm$ .008 & -.006 & -.009 \\
                           & 50\% & .803 $\pm$ .006 & .496 $\pm$ .024 & .796 $\pm$ .006 & .448 $\pm$ .018 & -.007 & -.048 \\
\hline
\multirow{2}{*}{Flex-MoE} & 20\% & .737 $\pm$ .019 & .452 $\pm$ .030 & .730 $\pm$ .011 & .405 $\pm$ .015 & -.007 & -.046 \\
                           & 50\% & .707 $\pm$ .007 & .461 $\pm$ .005 & .678 $\pm$ .009 & .341 $\pm$ .016 & -.028 & -.120 \\
\hline
\multirow{2}{*}{ShaSpec}  & 20\% & .860 $\pm$ .003 & .630 $\pm$ .004 & .851 $\pm$ .001 & .622 $\pm$ .002 & -.008 & -.008 \\
                           & 50\% & .846 $\pm$ .004 & .614 $\pm$ .002 & .825 $\pm$ .002 & .598 $\pm$ .001 & -.021 & -.015 \\
\toprule[1.pt]
\end{tabular}
\end{table*}

\begin{table*}[!t]
\centering
		\renewcommand{\arraystretch}{1.1}
    \small
\setlength{\tabcolsep}{7pt}% Adjust column spacing
\caption{Effect of diffusion-based imputation on PTB-XL under within-modality missing. \textbf{Raw}: models trained on zero-filled corrupted input. \textbf{Diffusion}: models trained on diffusion-reconstructed input. $\Delta$ denotes the performance change from Raw to Diffusion Imputed. \textbf{AUC}: AUROC; \textbf{F1}: Macro-F1. Three seeds; mean $\pm$ std.}
\label{tab:csdi_within}
\begin{tabular}{l c | cc | cc | cc}
\toprule[1.pt]
\multirow{2}{*}{\textbf{Model}} & \multirow{2}{*}{\textbf{Rate}} &
  \multicolumn{2}{c|}{\textbf{Raw}} &
  \multicolumn{2}{c|}{\textbf{Diffusion Imputed}} &
  \multicolumn{2}{c}{\textbf{$\Delta$}} \\
 & & AUC & F1 & AUC & F1 & AUC & F1 \\
\hline
\multirow{2}{*}{CLIMB}    & 20\% & .822 $\pm$ .003 & .519 $\pm$ .015 & .826 $\pm$ .003 & .526 $\pm$ .003 & +.004 & +.007 \\
                           & 50\% & .798 $\pm$ .004 & .457 $\pm$ .015 & .812 $\pm$ .005 & .487 $\pm$ .006 & +.014 & +.030 \\
\hline
\multirow{2}{*}{Flex-MoE} & 20\% & .736 $\pm$ .012 & .429 $\pm$ .025 & .760 $\pm$ .009 & .435 $\pm$ .026 & +.024 & +.006 \\
                           & 50\% & .706 $\pm$ .007 & .404 $\pm$ .015 & .744 $\pm$ .004 & .427 $\pm$ .013 & +.038 & +.023 \\
\hline
\multirow{2}{*}{ShaSpec}  & 20\% & .863 $\pm$ .004 & .634 $\pm$ .006 & .867 $\pm$ .001 & .636 $\pm$ .001 & +.004 & +.003 \\
                           & 50\% & .858 $\pm$ .004 & .630 $\pm$ .005 & .860 $\pm$ .001 & .631 $\pm$ .001 & +.001 & +.001 \\
\toprule[1.pt]
\end{tabular}
\end{table*}

\subsection{Degradation Analysis}
\label{sec:degradation}

Table~\ref{tab:family_degradation} retains the dataset, missingness mode, and rate as separate conditions. For each model and condition, the AUROC drop is clean minus missing AUROC, averaged over three seeds. Each family entry averages the drops of its member models. Positive drops indicate degradation and have the opposite sign to $\Delta_{\mathrm{ret}}$ in Table~\ref{tab:scale_summary}.

\begin{table}[!t]
\centering
\small
\setlength{\tabcolsep}{2pt}
\caption{Mean $\Delta$AUROC (clean $-$ missing) by model family, averaged over three seeds. CI: CLIMB and MIRA; SS: ShaSpec; MoE: Flex-MoE, FuseMoE, and Maestro. Larger values indicate greater performance drop.}
\label{tab:family_degradation}
\begin{tabular*}{\columnwidth}{@{\extracolsep{\fill}}llccc@{}}
\toprule
Dataset & Rate & CI & SS & MoE \\
\midrule
\multicolumn{5}{c}{\textbf{(a) Modality missing}} \\
\midrule
\multirow{2}{*}{Sleep-EDF}  & 20\% & .015 & .005 & .024 \\
                             & 50\% & .056 & .024 & .089 \\ \hline
\multirow{2}{*}{PTB-XL}     & 20\% & .011 & .005 & .024 \\
                             & 50\% & .026 & .019 & .058 \\ \hline
\multirow{2}{*}{Chapman}     & 20\% & .000 & .001 & .020 \\
                             & 50\% & .015 & .006 & .061 \\ \hline
\multirow{2}{*}{MIMIC-IV}   & 20\% & .024 & .019 & .029 \\
                             & 50\% & .067 & .041 & .072 \\ \hline
\multirow{2}{*}{Challenge}   & 20\% & .038 & $-$.035 & .053 \\
                             & 50\% & .038 & .003    & .113 \\ \hline
\multirow{2}{*}{PPG-DaLiA}  & 20\% & .008 & .006 & .024 \\
                             & 50\% & .031 & .023 & .067 \\ \hline
\multirow{2}{*}{WESAD}       & 20\% & .041 & .033 & .011 \\
                             & 50\% & .058 & .023 & .014 \\ \hline
\multirow{2}{*}{HAR-UP}     & 20\% & .007 & .007 & .034 \\
                             & 50\% & .026 & .027 & .096 \\ \hline
\multirow{2}{*}{CirCor}     & 20\% & .001 & .019 & .036 \\
                             & 50\% & .023 & .001 & .040 \\
\midrule
\multicolumn{5}{c}{\textbf{(b) Within-modality missing}} \\
\midrule
\multirow{2}{*}{Sleep-EDF}  & 20\% & .004 & .001   & .013 \\
                             & 50\% & .015 & .003   & .045 \\ \hline
\multirow{2}{*}{PTB-XL}     & 20\% & .005 & .002   & .019 \\
                             & 50\% & .026 & .007   & .054 \\ \hline
\multirow{2}{*}{Chapman}     & 20\% & .011 & .007   & .022 \\
                             & 50\% & .031 & .014   & .058 \\ \hline
\multirow{2}{*}{MIMIC-IV}   & 20\% & $-$.004 & .003 & .005 \\
                             & 50\% & .014    & .011 & .013 \\ \hline
\multirow{2}{*}{Challenge}   & 20\% & .016 & $-$.028 & .005 \\
                             & 50\% & .025 & $-$.018 & .020 \\ \hline
\multirow{2}{*}{PPG-DaLiA}  & 20\% & .004 & .008   & .010 \\
                             & 50\% & .019 & .020   & .037 \\ \hline
\multirow{2}{*}{WESAD}       & 20\% & .013 & .015   & $-$.004 \\
                             & 50\% & .054 & .035   & .018 \\ \hline
\multirow{2}{*}{HAR-UP}     & 20\% & .006 & .032   & .033 \\
                             & 50\% & .044 & .049   & .059 \\ \hline
\multirow{2}{*}{CirCor}     & 20\% & $-$.017 & .032 & .014 \\
                             & 50\% & .026    & .015 & .021 \\
\bottomrule
\end{tabular*}
\end{table}

Figure~\ref{fig:degradation}(left) summarizes AUROC degradation across all datasets and models; the family-level breakdown is given in Table~\ref{tab:family_degradation}. Figure~\ref{fig:degradation}(right) shows Flex-MoE on PPG-DaLiA. Macro-F1 drops more sharply than AUROC under both conditions. Modality missing produces the larger decline, and the increase from 20\% to 50\% is steeper for modality missing.

\subsection{PTB-XL Case Study: Diffusion Imputation}
\label{sec:imputation}
We evaluate an impute-then-predict strategy on PTB-XL. A conditional score-based diffusion model~\citep{ho2020ddpm,tashiro2021csdi} is trained at native ECG resolution ($T{=}1000$, 50 diffusion steps) conditioned on observed time points within each lead, with no cross-lead conditioning. The imputed signals are aggregated four-to-one to the $T{=}250$ input used by the benchmark classifiers, which then train on the aligned signals. Results are reported in Tables~\ref{tab:csdi_modality}--\ref{tab:csdi_within}.

\paragraph{Imputer Training and Offline Reconstruction.}
The imputer is trained independently for 20 epochs with 50 diffusion steps on contiguous masks covering 20\% or 50\% of each lead. It reconstructs the train, validation, and test splits offline; the outputs are aggregated from $T{=}1000$ to $T{=}250$, and classifiers load these arrays without additional online corruption. The modality-missing experiment applies the same imputer to entirely absent leads.

The diffusion imputer is trained with contiguous within-lead masks at
20\% and 50\% missingness and operates independently on each ECG lead.
For within-modality missing, it reconstructs the masked segments of an
observed lead. For modality missing, the same per-lead imputer is applied
to the entirely missing leads. In both settings, reconstruction is
performed offline at $T{=}1000$, followed by four-to-one aggregation to
the $T{=}250$ input used by the downstream classifiers.

Under modality missing, imputation lowers AUROC by 0.006 to 0.028 across the six model--rate settings. Macro-F1 drops by up to 0.120.
Under within-modality missing, mean AUROC improves for all three models. At 50\%, Flex-MoE gains 0.038 AUROC, CLIMB gains 0.014, and ShaSpec gains 0.001. Across all 18 within-modality comparisons (three models, two rates, and three seeds), imputation does not reduce AUROC relative to raw missing.

\section{Conclusion}
\label{sec:conclusion}

We introduce \method, a benchmark for multimodal fusion under incomplete physiological inputs. Across 9 datasets, 6 models, and 810 runs, the results show that robustness depends on the interaction between fusion design, data structure, and training exposure. Channel-independent models are often stable under whole-channel loss, while shared-specific models retain performance under contiguous gaps in homogeneous signals. The capacity-matched and transfer controls support these observations under narrower budgets and a shifted acquisition setting. A PTB-XL case study further suggests that diffusion imputation can help with within-channel gaps, while its benefit is limited when an entire input is absent. These findings support dataset-aware evaluation of missingness tolerance and cautious selection of fusion mechanisms for incomplete-on-arrival data.

\section*{Limitations}
\label{sec:limitations}
Our injected masks are controlled stress tests. They provide reproducible comparisons while preserving the native incomplete records in datasets such as MIMIC-IV and CirCor, but the benchmark does not estimate a clinical incidence rate or cover every natural missingness mechanism. Source-level correlated failures and a matched-rate comparison between native and injected missingness remain open directions. The benchmark evaluates fusion over physiological signals and pre-extracted image and text representations; it does not assess end-to-end robustness of raw image or language encoders. The cross-site analysis covers one PTB-XL to Chapman-Shaoxing transfer direction, and the imputation study is limited to PTB-XL. Finally, the six baselines retain model-specific adapters and training implementations, so the family comparisons describe the evaluated representatives within the shared data and evaluation protocol.

% -------------------------------------------------------

\bibliographystyle{IEEEtran}
\bibliography{references}

\end{document}